\documentclass[10pt,letterpaper,twocolumn]{article}
\usepackage[latin9]{inputenc}
\usepackage{color}
\usepackage{array}
\usepackage{verbatim}
\usepackage{float}
\usepackage{multirow}
\usepackage{amsmath}
\usepackage{amssymb}
\usepackage{graphicx}
\usepackage[unicode=true,
 bookmarks=false,
 breaklinks=true,pdfborder={0 0 1},backref=section,colorlinks=false]
 {hyperref}
\hypersetup{
 pagebackref=true,letterpaper=true,colorlinks}
\usepackage{breakurl}

\makeatletter

\providecommand{\tabularnewline}{\\}

\usepackage{iccv}
\usepackage{times}
\usepackage{epsfig}
\usepackage{array}
\usepackage{colortbl}
\usepackage{multirow}
\newcommand{\todo}[1]{{\color{red}{#1}}}

\iccvfinalcopy % *** Uncomment this line for the final submission

\def\iccvPaperID{2680} % *** Enter the ICCV Paper ID here

\ificcvfinal\fi

\makeatother

\begin{document}
%%%%%%%%% TITLE

\title{A Spatiotemporal Oriented Energy Network for Dynamic Texture Recognition}

\author{Isma Hadji\\
	York University, Toronto\\
	{\tt\small hadjisma@cse.yorku.ca}
	\and
	Richard P. Wildes\\
	York University, Toronto\\
	{\tt\small wildes@cse.yorku.ca}
}

\maketitle
%\todo{
\thispagestyle{empty}
\begin{abstract}
This paper presents a novel hierarchical spatiotemporal orientation
representation for spacetime image analysis. It is designed to combine
the benefits of the multilayer architecture of ConvNets and a more
controlled approach to spacetime analysis. A distinguishing aspect
of the approach is that unlike most contemporary convolutional networks
no learning is involved; rather, all design decisions are specified
analytically with theoretical motivations. This approach makes it
possible to understand what information is being extracted at each
stage and layer of processing as well as to minimize heuristic choices
in design. Another key aspect of the network is its recurrent nature,
whereby the output of each layer of processing feeds back to the input.
To keep the network size manageable across layers, a novel cross-channel
feature pooling is proposed. The multilayer architecture that results
systematically reveals hierarchical image structure in terms of multiscale,
multiorientation properties of visual spacetime. To illustrate its
utility, the network has been applied to the task of dynamic texture
recognition. Empirical evaluation on multiple standard datasets shows
that it sets a new state-of-the-art. %}
\vspace{-15pt}
\end{abstract}
%%%%%%%%% BODY TEXT
%------------------------------------------------------------------------- 
\section{Introduction\label{sec:intro}}

Hierarchical representations play an important role in computer vision \cite{Rodriguez2015}. The challenge of extracting
useful information (\eg objects, materials and environmental layout)
from images has led to incremental recovery of progressively more abstracted
representations. Convolutional networks (ConvNets) provide an interesting contemporary example of this paradigm yielding state-of-the-art results on a range of classification and
regression tasks, \eg \cite{Krizhevsky2012,Tran2015,Szegedy2013,Eigen2015}.
While such learning-based approaches show remarkable performance,
they typically rely on massive amounts of training data and the exact
nature of their representations often remains unclear. Deeper theoretical
understanding should lessen dependence on data-driven design, which
is especially important when training data is limited. In complement, the present work explores a more controlled approach to network realization. A small vocabulary of theory motivated, analytically
defined filtering operations are repeatedly cascaded to yield hierarchical
representations of input imagery. Although the same operations are
applied repeatedly, different information is extracted at each layer
as the input changes due to the previous layer's operations. Since
the primitive operations are specified analytically, they do not require
training data and the resulting representations are readily interpretable.
Further, the network yields state-of-the-art results on the important
and challenging task of dynamic texture recognition.

Previous work pursuing controlled approaches to hierarchical network
realization variously relied on biological inspiration and analytic principles.
Most biologically-based approaches mimic the multilayered architecture
of the visual cortex with cascades of simple and complex cells \cite{Riesenhuber99,MutchLowe2006,Jhuang2007}.
Typically, these approaches still use learning and leave open questions regarding the theoretical basis of their designs.
More theoretically driven approaches typically focus on network architecture
optimization, \eg the number of filters/layer or the number of learned
weights \cite{LeCun1998,Feng2015}. Other work made use of predetermined
filters at all layers, as learned via PCA \cite{PCAnet2015} or over
a basis of 2D derivatives \cite{Jacobsen16}.
Two commonalities appear across these efforts. First, they address
only a single aspect of their architecture, while assuming all others
are fixed. Second, they rely on learning in optimization. More closely
related to the present work is ScatNet \cite{Bruna2013}. This network
is rigorously defined to increase invariance to signal deformations
via hierarchical convolution with filters of different frequency
tunings. In its restricted application to 2D spatial images, ScatNet
bases its design on optimization with respect to 2D invariances. In
contrast, the present approach considers 3D spacetime images,
which leads to a spatiotemporal orientation analysis for extracting varying
dynamic signal properties across levels, including invariance maximization
via a multiscale network. Moreover, while the present approach
analytically specifies the number of filters/layer, ScatNet's choice
of wavelets limits it from analytically specifying the number of filters/layer,
which instead are chosen empirically.

Dynamic texture recognition also has received much attention. Here,
features used can be largely categorized as learned vs. hand-crafted.
Those in the former category rely on either autoregressive \cite{Szummer1996}
or Linear Dynamical Systems (LDS) \cite{saisanDWS01}. Recent trends
typically rely on LDS with dictionary learning \cite{Mumtaz2013,Harandi2013,QuanHJ15}.
The main downside of such approaches comes from their inability to
represent patterns beyond their training data. 
In contrast, hand-crafted approaches to dynamic texture analysis typically
eschew modeling the underlying dynamical system in favor of more directly
encoding the spacetime signal with an attempt to balance discriminability
and generalizability. Such approaches can be subcategorized according
to treating each frame as a static texture \cite{Ghanem2010,Yang2016}, 3D Local Binary
Patterns \cite{Zhao2007,Ren2013}, reliance on optical flow \cite{Peteri2005,Lu2005}
and building more directly on spatiotemporal filtering \cite{KostaTexture,Xu2011,Ji2013,Xu2012,Dubois2013}.
This last approach is most akin to the present work, as it also has at its core
the application of spatiotemporal filters to data streams. In contrast,
the proposed approach exploits repeated application of such filters
and combination of their outputs at each layer to extract progressively
more abstract information.
\textbf{Contributions.} In light of previous research, the contributions
of this paper are as follows. \textbf{1)} A novel processing network,
based on a repeated spatiotemporal oriented energy analysis, is presented where
the layers interact via a recurrent connection so the output feeds
back to the input. The resulting multilayer architecture systematically
reveals the hierarchical structure in its input. Exposed properties
progress from simple and local to abstract and global across layers,
but are always interpretable in terms of the network's explicit design.
\textbf{2)} At every layer, extracted feature maps are combined via
cross-channel pooling to yield 
 coherent groups based on the employed filters. This innovation constrains
the representation dimensionality while maintaining interpretability
and high discriminating power. \textbf{3)} Every
stage of processing in the network is designed based on theoretical
considerations. Ties to biological modeling are also established.
This design approach removes the need for a learning phase, which
is not always feasible, \eg, when confronted with modest datasets.
\textbf{4)} The resulting network is used
as a novel approach to representation and recognition of dynamic
textures (DTs). In empirical evaluation on standard datasets, the
system achieves superior performance to virtually all current alternative
DT recognition approaches.
Code is available at \todo{https://github.com/hadjisma/soe-net}.\vspace{-5pt}

%------------------------------------------------------------------------- 
\vspace{-1pt}

\section{Technical approach\label{sec:Technical-Approach}\vspace{-5pt}
}

\begin{figure}
\begin{centering}
\includegraphics[width=0.85\columnwidth,height=0.2\textheight]{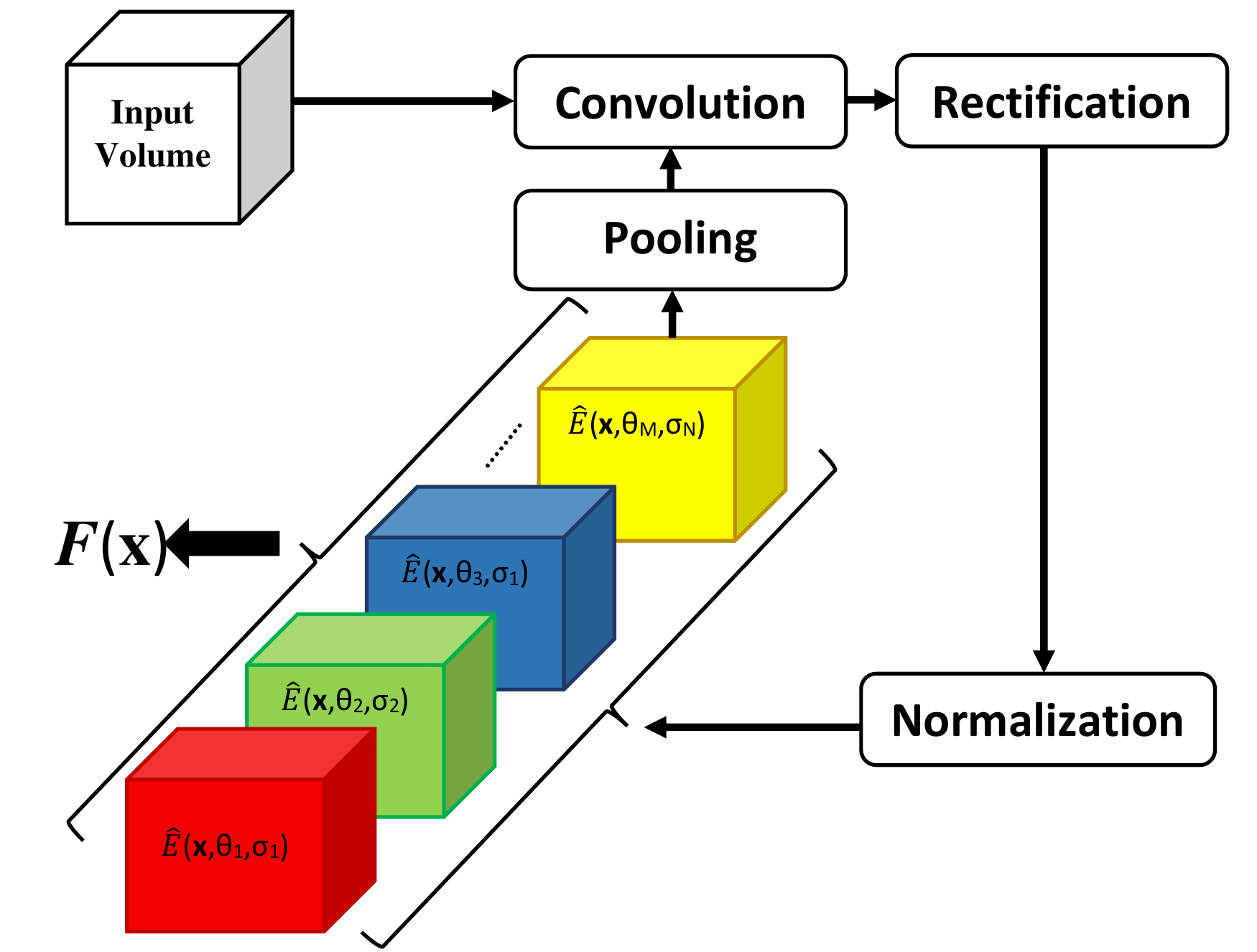} 
\par\end{centering}
\caption{\label{fig:architecture}Overview of the \textbf{SOE-Net} Architecture.
The same set of operations are repeatedly applied via a recurrent
connection; however, different information, $\boldsymbol{F}$, is
extracted at each pass as the input changed due to the operations
of the previous pass.} 
\vspace{-15pt}
\end{figure}

The proposed network architecture is designed to capture spatiotemporal
image structure across multiple layers of processing 
 as shown in Fig.$\,$\ref{fig:architecture}. The input to the system
is a three-dimensional, $\mathbf{x}=(x,y,t)^{\top}$, spacetime volume,
$V(\mathbf{x})$, 
and the output is a volume of feature maps, $\boldsymbol{F}(\mathbf{x})$,
that capture the spatiotemporal structure. Each processing layer,
$\mathcal{L}_{k}$, is comprised of a sequence of four stages: convolution,
rectification, normalization and pooling. 
A key novelty of the approach is the repeated filtering, whereby the final processing
stage of each layer (pooling) feeds back to the initial processing
stage (convolution) to yield a subsequent layer of processing, $\mathcal{L}_{k+1}$.
The entire process is repeated $K$ times, after which the energy
remaining in the signal about to be fed back through has essentially
vanished.
 The final output of the network, $\boldsymbol{F}(\mathbf{x})$, is
the set of feature maps extracted at the $K^{th}$ layer. Each layer in the network corresponds to Spatiotemporal
Oriented Energy filtering; therefore, the network is dubbed \textbf{SOE-Net}.

\vspace{-2pt}

\subsection{Repeated filtering\label{subsec:The-Recurrence}\vspace{-2pt}
}

Key to the success of multilayer architectures (\eg ConvNets) is
their ability to extract progressively more abstract attributes of
their input at each successive layer and thereby yield a powerful
data representation. As a simple example, a degree of shift-invariance
emerges via linear filtering, followed by nonlinear rectification
and pooling: The exact position of the extracted features becomes
immaterial across the pooling support. A similar interpretation becomes
more subtle, however, as the entire filter-rectify-pool block of operations
is repeated, especially when using learned filters.

\begin{figure*}
\begin{centering}
\includegraphics[width=0.9\textwidth,height=0.22\textheight]{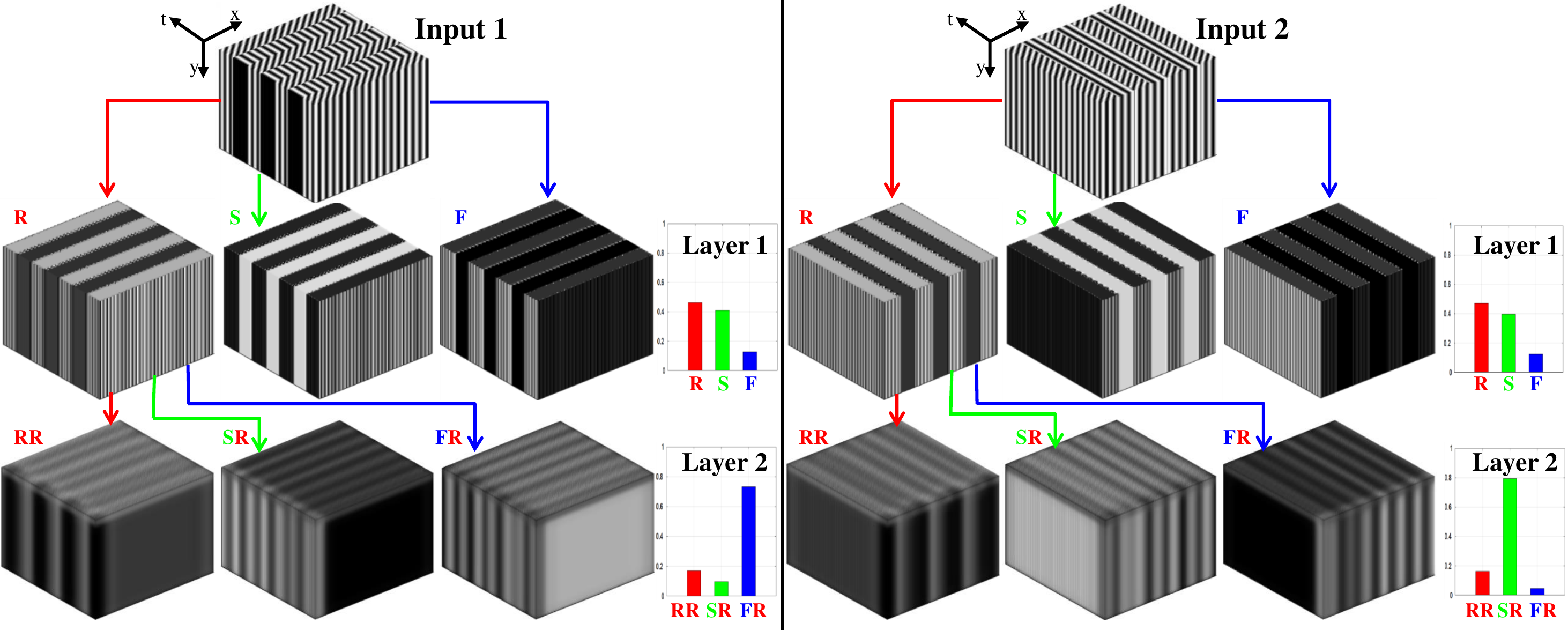} 
\par\end{centering}
\caption{\label{introfigure}Emergence of Abstract Features via Repeated Filtering.
(Left) Input synthetic sinusoidal pattern alternately moves right (orientation
along $x$-$t$ diagonal) and stays static (orientation parallel to
time axis). 
 (Right) Same pattern moving right behind a static picket
fence with same spatial pattern (\ie background motion viewed between
the spaces in a stationary picket fence). \textbf{SOE-Net } is used
with filters tuned to \textbf{R}ightward motion, \textbf{S}tatic (no
motion) and \textbf{F}lickering (pure temporal change). \textbf{L}ayer\textbf{ 1}
captures the local rightwardly moving and static portions; but not
the more abstract alternating temporal move-stop pattern of the left
input, nor the fact that the right input maintains the same spatially
interleaved moving and static stripes. Indeed, measurements aggregated
over the volumes (shown as histograms on the right) are the same at
this layer. 
The difference is revealed at \textbf{L2} after applying the same
filters on the \textbf{R} response of \textbf{L1}: The move-stop behavior
of the left texture becomes explicit and yields a large \textbf{F}
response. 
In contrast, the constant rightward motion and picket fence of the
right texture yield a large \textbf{S} response.
 In practice, more directional filters are applied at each layer;
see Sec.~\ref{sub:convolution}.}
\vspace{-15pt}
\end{figure*}

\begin{figure}
\begin{centering}
\includegraphics[width=0.85\columnwidth,height=0.2\textheight]{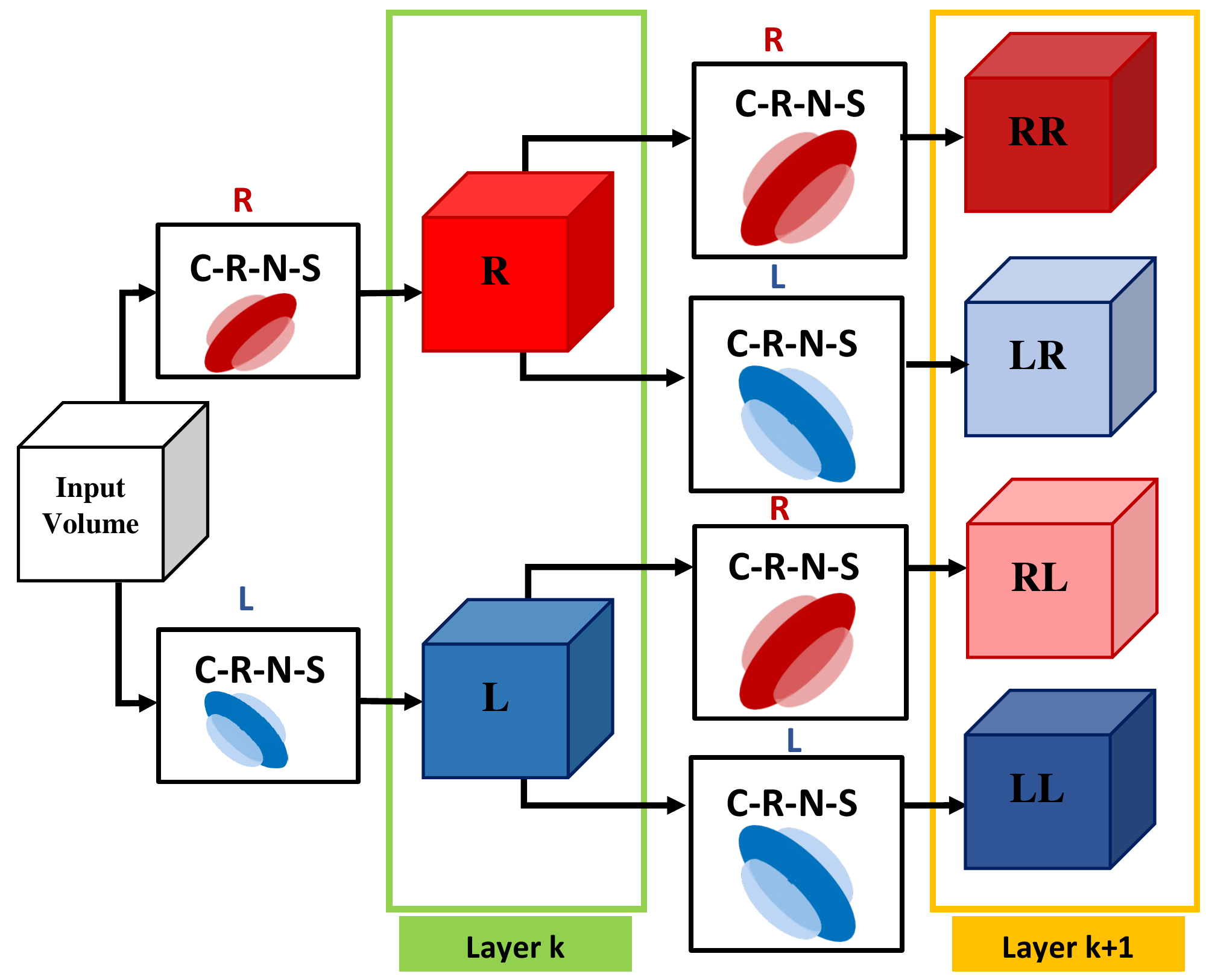} 
\par\end{centering}
\caption{\label{fig:recurrence}Unfolding the \textbf{SOE-Net} Recurrent Connection.
Local spatiotemporal features at various orientations are extracted
with an initial processing layer, $\mathcal{L}_{k}$. $\sf{C}$-$\sf{R}$-$\sf{N}$-$\sf{S}$
indicate Convolution, Rectification, Normalization and Spatiotemporal
pooling, as detailed in Secs.~\ref{sub:convolution}, \ref{sub:rectification},
\ref{sub:normalization} and \ref{sub:pooling}, resp., while R and
L indicate rightward vs. leftward filtered data, resp., with symbol
strings (\eg LR) indicating multiple filterings. A network
with only 2 filters (\ie 2 orientations) is shown
for illustration. Each of the feature maps
at layer $\mathcal{L}_{k}$ is treated as a new separate signal and
fed back to layer $\mathcal{L}_{k+1}$ to be convolved with the same
set of filters but at a different effective resolution due to
spatiotemporal pooling.}
\vspace{-20pt}
\end{figure}

Consider the simple example of Fig.~\ref{introfigure}. The left
side depicts a sinusoidalpattern alternately moving rightward and staying static across time.
On the right, the same 
pattern is moving rightward behind a similarly textured static picket
fence. An initial stage of directional (motion) filtering cannot extract
the difference in the overall dynamic behavior of the two patterns.
However, further directional filtering that operates on the output
of the initial layer can detect the overall patterns and thereby allows to make the distinction
between the two different dynamic textures. More generally, this example
shows how powerful features can emerge across multiple layers of processing:
The exposed properties progress from simple and local to abstract
and global.
Motivated by these observations, the proposed SOE-Net is designed
to extract progressively more abstract representations of the input
signal at each layer, while maintaining interpretability. To achieve
these ends, repeated filtering is employed via a recurrent connection, whereby the output
of layer $\mathcal{L}_{k}$, denoted as $\boldsymbol{L}_{k}$, feeds
back to the initial convolutional stage (\ie convolution with the
same filter set) to yield a subsequent layer of processing, $\mathcal{L}_{k+1}$.
Since the processing at each layer is defined precisely (see following
subsections), it will be interpretable. Also, since it is applied
repeatedly, abstraction emerges. This process is depicted in Fig.
\ref{fig:recurrence} with an unfolded recurrence, and symbolized
as\vspace{-8pt}
\begin{equation}
%\vspace{-5pt}
\boldsymbol{L}_{k+1}=\mathcal{L}_{k+1}\left(\boldsymbol{L}_{k}\right),\vspace{-1pt}\vspace{-5pt}\label{eq:recurrence}
\end{equation}
with $\boldsymbol{L}_{1}=\mathcal{L}_{1}\left(V(\mathbf{x})\right)$.
Interestingly, biological models have advocated that similar processing
(termed $Filter\rightarrow Rectify\rightarrow Filter$) takes place
in visual cortex to deal with higher-order image structures \cite{baker2001}.

%[H]

\subsection{Convolution\label{sub:convolution}}\vspace{-5pt}

Convolution serves to make local measurements revealing salient properties
of 
its input. Given a temporal sequence of images, local spatiotemporal
orientation is of fundamental descriptive power, as it captures the
$1^{st}$-order correlation structure of the data irrespective of
the underlying visual phenomena \cite{Adelson91,wildes2000,KostaTexture}.
Also, such measures provide a uniform way to capture spatial appearance
and dynamic properties of the underlying structure. 
Spatial patterns (\eg static texture) are captured as the filters
are applied within the image plane. Dynamic attributes of the pattern
are obtained by filtering at orientations extending into time. 

Based on the above theoretical motivations, in this work convolution
is designed to extract local measurements of multiscale, spatiotemporal
orientation. Still, having committed to convolution that captures
spatiotemporal orientation, a variety of options exist for specifying
an exact set of filters, \eg. Gabor, lognormal, wavelet or Gaussian
derivatives. Here, Gaussian derivative filters are selected for two
main reasons, \cf\cite{Freeman1991}. First, for any given order
of Gaussian derivative, a set of basis filters that allow synthesis
of the response at an arbitrary orientation can be specified. This
property makes it possible to set the exact number of filters used
at each layer on a theoretical basis, rather than experiments or learning.
Second, these filters are separable, which provides efficient implementation.
In particular, 3D Gaussian $3^{rd}$- order derivative filters are
used, $G_{3D}^{(3)}(\theta_{i},\sigma_{j})$, 
with $\theta_{i}$ and $\sigma_{j}$ denoting the 3D filter orientation
and scale, resp. 
Given an input spacetime volume, $V(\mathbf{x})$, a set of output
volumes, $C(\mathbf{x};\theta_{i},\sigma_{j})$, are produced as 
\vspace{-5pt}
\begin{equation}
C(\mathbf{x};\theta_{i},\sigma_{j})=G_{3D}^{(3)}{\displaystyle (\theta_{i},\sigma_{j})*V(\mathbf{x})},\vspace{-5pt}\label{eq:convolution}
\end{equation}
with $*$ denoting convolution. Since $3^{rd}$- order Gaussian filter
are used, $M=10$ filters are required to span the space of orientations
\cite{Freeman1991}. The directions, $\theta$, are chosen to uniformly
sample 3D orientations as the normals to the faces of an icosahedron,
with antipodal directions identified. The number of scales, $\sigma$,
is determined by the size of the spacetime volume to be analyzed; details are provided in Sec.~\ref{subsec:Implementation-Details}.

\vspace{-3pt}

\subsection{Rectification\label{sub:rectification}\vspace{-3pt}
}

The output of the convolutional stage, $C(\mathbf{x};\theta_{i},\sigma_{j})$,
is comprised of positive and negative responses. Both indicate a contrast
change along the direction of the oriented filters and hence structure
along that direction. It is therefore interesting to keep the distinction
between the two responses. Also, in anticipation of the subsequent
pooling stage of processing, it is critical to perform some type of
rectification. Otherwise, the pooling across positive and negative
responses will attenuate the signal. By squaring the individual responses,
it is possible to consider the results in terms of spectral energy,
\eg \cite{wildes2000}. Based on these considerations, the present
approach makes use of a nonlinear operation that splits the signal
into two paths: The first path carries the positive responses, while
the second carries the negative responses, both of which are pointwise
squared to relate to spectral energy measurements to yield \vspace{-3pt}
\begin{equation}
\begin{array}{c}
E^{+}(\mathbf{x};\theta_{i},\sigma_{j})=\left(\max[C(\mathbf{x};{\displaystyle \theta_{i},\sigma_{j}),0]}\right)^{2}\\
E^{-}(\mathbf{x};\theta_{i},\sigma_{j})=\left(\min[C(\mathbf{x};{\displaystyle \theta_{i},\sigma_{j}),0]}\right)^{2}
\end{array}.\vspace{-3pt}\label{eq:rectification}
\end{equation}

Interestingly, biological findings suggest a model for cortical simple
cells that includes a nonlinearity in the form of two half wave rectifications
that treat the positive and negative outputs in two different paths
\cite{Heeger91}. Similarly, recent ConvNet analysis also revealed
that learned filters tend to form pairs with opposite phases \cite{SHANG2016}.

Beyond physical interpretation (signed energy) and relation to biology,
use of the squaring function at this stage is advantageous as it allows
for keeping track of the frequency content of the processed signal
(\ie doubling the maximum frequency present). This information plays
an important role in specifying the pooling stage parameters, detailed
below. 
Given that the following processing stages treat the two paths in
the exact same way, in the remainder of the paper the $+$ and $-$
superscripts will be suppressed when referring to energy measurements,
$E(\mathbf{x};\theta_{i},\sigma_{j})$. 

\subsection{Normalization\label{sub:normalization}}\vspace{-5pt}

Owing to the bandpass nature of the filtering used in the convolutional
stage, \eqref{eq:convolution}, the responses are invariant to overall
additive brightness offsets. Still, the responses remain sensitive
to multiplicative contrast variations, \ie the filter response increases
with local contrast independent of local orientation structure. Moreover,
the responses after rectification, \eqref{eq:rectification}, are
unbounded from above, which makes practical signal representation
challenging. Divisive normalization is a way to correct for these
difficulties. Significantly, it also serves to lessen statistical
dependencies present in natural images via signal whitening \cite{simoncelli2008}.
Therefore, the next stage operates via pointwise division of the rectified
measurements by the sum over all orientations to yield\vspace{-5pt}
\begin{equation}
\hat{E}(\mathbf{x};\theta_{i},\sigma_{j})=\frac{E(\mathbf{x};\theta_{i},\sigma_{j})}{\sum_{m=1}^{M}E(\mathbf{x};\theta_{m},\sigma_{j})+\epsilon}.\vspace{-5pt}\label{eq:normalization}
\end{equation}
Here, $\epsilon$ serves as a noise floor and to reduce numerical
instabilities. It is specified as the standard deviation of the energy
measurements across all orientations. Once again, it is interesting
to note that biological modeling of visual processing have employed
a similar divisive normalization, including use of a signal adaptive
saturation constant \cite{Heeger91}. Finally, note that features
captured at this layer, 
$\hat{E}(\mathbf{x};\theta_{i},\sigma_{j})$, are measures of Spatiotemporal
Oriented Energy \cite{wildes2000,KostaTexture}; therefore, the overall
network is named \textbf{SOE-Net}.

\vspace{-3pt}

\subsection{Pooling\label{sub:pooling}\vspace{-5pt}
}

Two pooling mechanisms are employed to achieve the desired level of
abstraction. First, spatiotemporal pooling is performed following
a frequency decreasing path. Thus, the same set of filters in the
convolution block operate on lower spatiotemporal frequencies at each
subsequent layer thereby revealing new information from the originally
input signal. Second, the feature maps extracted for each layer are
linearly combined, through cross-channel pooling, to capture more
complex structures at each layer while preventing the number of feature
maps from exponential increase.

\textbf{Spatiotemporal pooling.} 
Spatiotemporal pooling serves to aggregate normalized responses, \eqref{eq:normalization},
over spacetime. Aggregation provides for a degree of shift invariance
in its output, as the exact position of a response in the pooling
region is abstracted. 
In SOE-Net, spacetime pooling is implemented via low-pass filtering
with a 3D Gaussian, $G_{3D}(\gamma)$, followed by downsampling, $\downarrow_{\tau}(\cdot)$,
where $\gamma$ is the standard deviation and $\tau$ is the sampling
period, \vspace{-5pt}
\begin{equation}
\boldsymbol{S}_{k}(\mathbf{x};\theta_{i},\sigma_{j})=\downarrow_{\tau}\left(G_{3D}(\gamma)*\hat{E}(\mathbf{x};\theta_{i},\sigma_{j})\right).\vspace{-3pt}\vspace{-5pt}\label{eq:st-pooling}
\end{equation}

Here, a key question is how to specify the pooling parameters, $\gamma$
and $\tau$? Typical ConvNets rely on heuristic choices, as their
learning based approach does not yield enough theoretical insight
for a formal answer. In contrast, since the signal properties of the
present architecture have been specified precisely, these parameters
can be specified analytically. First, applying a $3^{rd}$- order Gaussian
derivative filter, $G_{3D}^{(3)}$, with zero mean and standard deviation,
$\sigma$, in the convolution stage greatly attenuates frequencies
$\omega_{c}>\frac{3\sqrt{3}}{\sigma}$ (\ie a factor of $\approx$~3
beyond the central frequency). However, given the frequency doubling
effect from the squaring in the rectification stage, consideration
must instead be given to $\eta=2\omega_{c}$ as the effective cut-off
frequency. Correspondingly, it is appropriate to select the low-pass
filter of the pooling stage with a cut-off frequency $\omega_{l}=\alpha\eta$
, with $0<\alpha<1$ to ensure operations on lower spatiotemporal
frequencies at each layer. This implies taking $\gamma=\frac{3}{\alpha\eta}$
(\ie approximating the cut off frequency to be $\approx$~3 standard
deviations away from the central frequency).

To avoid aliasing in downsampling, the sampling theorem is used such
that $\omega_{s}>2\alpha\eta$. Correspondingly, the sampling period
is $\tau=\beta\frac{2\pi}{\omega_{s}},0<\beta<1$. In implementation,
a conservative $\beta=0.5$ is employed. Notably, use of low-pass
filtering with decreasing frequencies guarantees an energy decay from
layer $\mathcal{L}_{k}$ to layer $\mathcal{L}_{k+1}$.

\begin{figure}
\begin{centering}
\includegraphics[width=0.9\columnwidth]{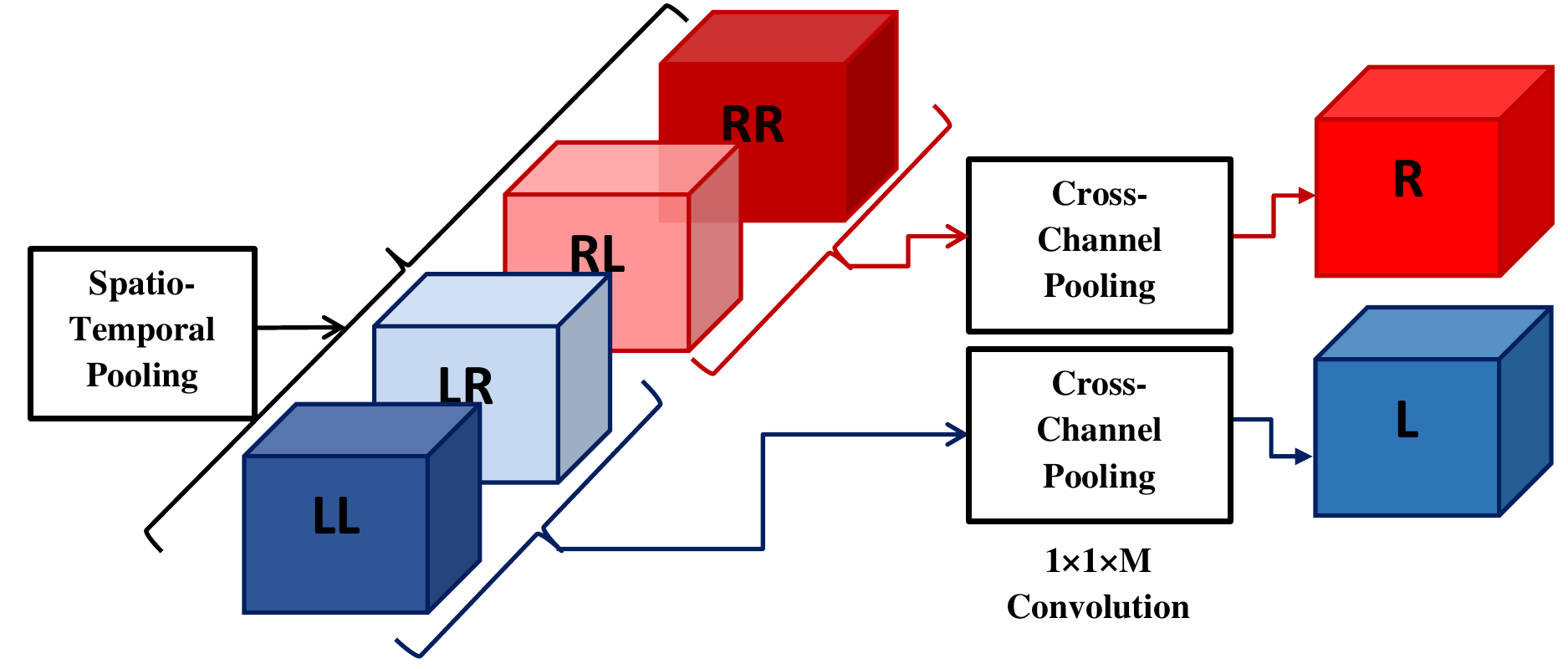} 
\par\end{centering}
\caption{\label{fig:channel_wise_pooling}Overview of the Proposed Cross-Channel
Pooling.}
\vspace{-15pt}
\end{figure}

\textbf{Cross-channel pooling.} Cross-channel pooling serves to aggregate
pooled responses, \eqref{eq:st-pooling}, across feature maps. Figure~\ref{fig:recurrence}
unfolds the recurrence to illustrate how once all features maps from
layer $\mathcal{L}_{k}$ have gone separately through the recurrence,
the number of feature maps, $\boldsymbol{S}_{k}$, for each scale,
$\sigma_{j}$, used in the convolution block is $M^{k}$. This situation
is unsatisfactory for two reasons. First, it fails to capture the
emergence of common attributes across feature maps that result from
applying the same filter orientation, $\theta_{i}$, to the inputs
from $\mathcal{L}_{k-1}$. Second, there is potential for explosion
of the representation's size as many layers are cascaded. Both of
these concerns can be dealt with by pooling across all feature maps
from $\mathcal{L}_{k}$ that result from filtering with a common orientation,
as illustrated in Fig.~\ref{fig:channel_wise_pooling}.

To formalize pooling across feature maps, recall that beyond the very
first layer, each $\boldsymbol{S}_{k}$ derives from input that was
itself parameterized by an orientation, $\theta_{m}^{k-1}$, from
filtering at the previous layer, $\mathcal{L}_{k-1}$. This dependence
is now captured explicitly by extending its parameterization to $\boldsymbol{S}_{k}(\mathbf{x};\theta_{i},\sigma_{j},\theta_{m}^{k-1})$.
This extension allows the desired cross-channel pooling to produce
the final output of $\mathcal{L}_{k}$ as \vspace{-6pt}
\begin{equation}
\boldsymbol{L}_{k}(\mathbf{x};\theta_{i},\sigma_{j})=\frac{1}{M}\sum_{m=1}^{M}\boldsymbol{S}_{k}(\mathbf{x};\theta_{i},\sigma_{j},\theta_{m}^{k-1}).\vspace{-5pt}\label{eq:C-pooling}
\end{equation}
In implementation this operation is realized as a $1\times1\times M$
convolution with an averaging kernel. Note that the summation holds
vacuously at the very first layer of processing.\vspace{-10pt}

Theoretically, \eqref{eq:C-pooling} can be seen as an oriented energy
generalization of the construction for derivatives of multivalued
images \cf\cite{dizenzo1986,sapiro1996,rubner1996}, but is novel
in the current context of oriented energy processing. In particular,
given a multivalued function \textbf{$\boldsymbol{f}(\mathbf{x})=[f_{1}(\mathbf{x}),...,f_{M}(\mathbf{x})]^{T}$
}with $\mathbf{x}=(x_{1},...,x_{N})^{T}$ and defining ${\sf G}=\sum_{m=1}^{M}\nabla f_{m}(\mathbf{x})\nabla^{T}f_{m}(\mathbf{x})$,
a measure of change along $\mathbf{\hat{v}}$ is given as $\mathbf{\hat{v}}^{T}{\sf G}\mathbf{\hat{v}}$.
Note that letting $\mathbf{\hat{v}}=\mathbf{\hat{e}}_{i}$, with $\mathbf{\hat{e}}_{i}$
the unit vector along $x_{i}$, yields $\mathbf{\hat{e}}_{i}^{T}{\sf G}\mathbf{\hat{e}}_{i}=\sum_{m=1}^{M}(\frac{\partial f_{m}}{\partial x_{i}})^{2}$.
Comparatively, in \eqref{eq:C-pooling} the multivalued function is
just the set of feature maps obtained from a previous layer, $\mathcal{L}_{k-1}$,
that has been differentially filtered and squared in the convolution
and rectification blocks, resp. Thus, $\boldsymbol{L}_{k}(\mathbf{x};\theta_{i},\sigma_{j})$
captures the energy along direction $\theta_{i}$ taken across the
multivalued $\boldsymbol{S}_{k}(\mathbf{x};\theta_{i},\sigma_{j},\theta_{m}^{k-1})$.

Conceptually, the cross-channel pooling, \eqref{eq:C-pooling}, produces
a set of $M$ feature maps (one for each filter orientation, $\theta$)
that capture the amount of structure present in the previous layer
at that orientation, irrespective of the source (\ie irrespective
of a particular feature map at the previous layer). This approach
thereby yields immediate insight into the nature of the representation,
in contrast to alternative approaches that rely on random cross-channel
combinations, \eg \cite{minlin14}. Reflecting back on the motivating
examples, Fig.$\,$\ref{introfigure}, the proposed processing was
used to generate the feature maps, yielding exactly the desired abstractions.

\vspace{-3pt}

\subsection{Dynamic texture recognition\label{sub:Recognizing-Dynamic-Textures}}

\vspace{-3pt}

The SOE-Net representation, $\boldsymbol{F}(\mathbf{x};\theta_{i},\sigma_{j})$,
extracted after the normalization block of the $K^{th}$ layer, provides
a rich hierarchical description of visual spacetime in terms of multiscale
spatiotemporal orientation. As such, it has potential to serve as
the representational substrate for a wide variety of spacetime image
understanding tasks (\eg segmentation, tracking and recognition).
As an illustrative example, dynamic texture (DT) recognition is considered.
This task has been considered for two main reasons. First, DTs are
pervasive phenomena in the visual world. Second, the ability to recognize
these patterns provides the basis for a number of further applications,
including video indexing, scene recognition and change detection in
surveillance and monitoring.

For the specific application of dynamic texture recognition, each
spacetime volume, $V(\mathbf{x})$, to be recognized is taken to contain
a single pattern. Therefore, the pointwise extracted feature maps,
$\boldsymbol{F}(\mathbf{x};\theta_{i},\sigma_{j})$, can be aggregated
over a region, $\Omega$, that covers the entire spacetime texture
sample to be classified to yield a global feature vector,\vspace{-5pt}
\begin{equation}
\mathcal{F}(\theta_{i},\sigma_{j})=\sum_{\mathbf{x}\in\Omega}\boldsymbol{F}(\mathbf{x};\theta_{i},\sigma_{j}).\vspace{-6pt}\label{eq:texture}
\end{equation}
The resulting global feature vector is $l_{2}$ normalized to yield
the overall descriptor $\hat{\mathcal{F}}(\theta_{i},\sigma_{j})$. To compare the feature distributions of an input query to database entries, the Bhattacharyya coefficient is used, as it provides a principled
way to compare two normalized distributions kept as histograms.

While previous work made use of spatiotemporal oriented energies for
dynamic textures \cite{KostaTexture}, it differs from the current
work in five significant ways. First and foremost, previous work did
not use repeated filtering, \eqref{eq:recurrence}, for hierarchical
abstraction of image structure. Second, it marginalized appearance
information so that it did not capture purely spatial structure and
therefore was less able than the current approach to capitalize on
both appearance and dynamics. Third, it did not separate the opposite
phase responses in rectification, \eqref{eq:rectification}. Fourth,
it did not employ multiple scales, $\sigma$, \eqref{eq:convolution}.
Fifth, no cross-channel pooling was involved in previous work, while
it has a key role in the current work, \eqref{eq:C-pooling}.

\vspace{-3pt}

\subsection{Implementation details\label{subsec:Implementation-Details}}

\vspace{-3pt}

Convolution with the Gaussian $3^{rd}$ derivatives, \eqref{eq:convolution},
is realized with separable 13-tap filters, with $M=10$ orientations/scale and $\sigma=1$.
Multiscale filtering is realized by applying the same filters across
levels of a Gaussian pyramid, with factor of 2 subsampling between
levels. The number of scales, $|\sigma|$, is chosen automatically to avoid undue
border effects depending on the size of the input spacetime volume:
The coarsest scale is the last such that the filter can fit entirely
within the corresponding pyramid level. Unless otherwise noted, this
constraint yields $|\sigma|=2$ for all datasets, except Dyntex++
where $|\sigma|=1$ due to the very small size of its videos $(50\times50\times50)$.
Similarly, the number of iterations over the network, \eqref{eq:recurrence},
\ie the number of layers, $K$, is stopped when the spatiotemporal
support of the signals to be fed back through the recurrence is less
than or equal to the filter size; see Sec.~\ref{sec:empirical} for
specifics by dataset. 
Due to the cross-channel pooling, \eqref{eq:C-pooling}, the output
of the convolution block starting from layer 2 is always $M^{2}\times|\sigma|$
feature maps. Also, because the proposed rectification strategy, \eqref{eq:rectification},
splits the results of each convolutional block into $2$ paths, the
effective number of feature maps after the rectification block is
doubled at each layer. Therefore, the dimension, $D_{K}$, of feature
vectors extracted after the normalization block of the $K^{th}$ layer
is $D_{K}=M^{2}\times|\sigma|\times2^{K}$. \vspace{-5pt}

 %------------------------------------------------------------------------- 

\section{Empirical evaluation\label{sec:empirical}\vspace{-3pt}
}

SOE-Net is evaluated according to standard protocols on the two most
recent dynamic texture datasets, YUVL \cite{KostaTexture}
and Dyntex \cite{PeteriFH10}. YUVL is the larger with 610 sequences
grouped into 5 classes (YUVL1) or by further subdividing 3 of the
initial classes into 6 subclasses (YUVL2), but only using 509 sequences
from the full set; see \cite{KostaTexture}. Here, a third organization
is introduced (YUVL3), where 2 classes neglected in YUVL2 are reinstated
to use the entire set divided into 8 classes. Dyntex is used in 5
main variations: Dyntex-35 has 35 classes with 10 sequences/class
\cite{Zhao2006}; Dyntex++ has 36 classes with 100 sequences/class
\cite{Ghanem2010}; Alpha, Beta and Gamma have 60, 162 and 275 sequences, divided into 3, 10 and 10 classes, resp \cite{Dubois2013}. The same protocol is used for all datasets. Feature vectors from
the last layer of SOE-Net are input to a Nearest-Neighbor (NN) classifier
using the leave-one-out procedure \cite{saisanDWS01,KostaTexture,QuanHJ15}.
Although NN is not state-of-the-art as a classifier, it is appropriate
here where the goal is to highlight the discriminative power of the
features without obscuring performance via use of more sophisticated
classifiers. Still, for completeness and fair comparison
to previous work, results also are reported using Nearest Class Center
(NCC) and SVM classifiers in Sec.~\ref{subsec:Comparison-to-DT-SOA}.\vspace{-5pt}

\subsection{Component-wise validation\vspace{-3pt}
}

SOE-Net's theory based component specifications are now validated
empirically. Primarily, classification accuracy on YUVL is used for
this goal, as it is the largest available and is well organized according
to pattern dynamics.

\textbf{Convolution.} The theoretical basis for
use of multiscale, spatiotemporally oriented filters, in general,
and Gaussian derivative filters, in particular, was given in Sec.~\ref{sub:convolution}. Still, a choice remains regarding the order of the derivative used.
Table~\ref{G2-VS-G3-VS-G4} compares $2^{nd}$-,
$3^{rd}$- and $4^{th}$-order Gaussian derivatives, in terms of classification accuracy, while using a
single layer and scale of SOE-Net. The results show $G3$ as the best
performer. This choice provides the right balance between
tuning specificity and the numerical stability of relatively
low-order filtering. In light of these observations, all subsequent
results rely on $3^{rd}$-order Gaussian derivatives.

\vspace{-7pt}
\begin{table}[H]
	\resizebox{\columnwidth}{!}{
		
		\begin{tabular}{c|c|c|c|c|c|c|c|c|c}
			\hline 
			& \multicolumn{3}{c|}{YUVL1} & \multicolumn{3}{c|}{YUVL2} & \multicolumn{3}{c}{YUVL3}\tabularnewline
			\hline 
			\multirow{2}{*}{SOE-Net (3D)} & G2  & G3  & G4  & G2  & G3  & G4  & G2  & G3  & G4\tabularnewline
			\cline{2-10} 
			& 83.3  & \textbf{91.1}  & 89.8  & 85.1  & \textbf{90.2}  & 90.1  & 74.1  & \textbf{84.6}  & 82.9\tabularnewline
			\hline 
		\end{tabular}
		
	}
	
	\caption{\label{G2-VS-G3-VS-G4}Comparison of $2^{nd}$, $3^{rd}$, $4^{th}$
		order Gaussian derivatives.}
	\vspace{-12pt}
\end{table}
Also, one could question the benefit of 3D filtering that captures temporal information
in recognizing dynamic textures over 2D filtering that captures only spatial appearance. Thus, a 2D version of SOE-Net is used as a baseline comparison. Further, a 2D state-of-the-art hand crafted network, originally proposed for 2D texture analysis, ScatNet \cite{Bruna2013} is compared. In both cases, 2D frames are supplied to the 2D networks. Table~\ref{3D-VS-2D} shows the decided advantage of
3D over 2D filtering for dynamic texture recognition.

\vspace{-8pt}
\begin{table}[H]
\begin{centering}
\resizebox{0.8\columnwidth}{0.03\textheight}{ 
\par\end{centering}
\begin{centering}
\begin{tabular}{c|c||c||c|c||c||c|c||c||c}
\hline 
 & \multicolumn{3}{c|}{YUVL1} & \multicolumn{3}{c|}{YUVL2} & \multicolumn{3}{c}{YUVL3}\tabularnewline
\hline 
ScatNet (2D) \cite{Bruna2013}  & \multicolumn{3}{c|}{68.7} & \multicolumn{3}{c|}{69.7} & \multicolumn{3}{c}{64.8}\tabularnewline
\hline 
SOE-NET (2D)  & \multicolumn{3}{c|}{64.0} & \multicolumn{3}{c|}{70.1} & \multicolumn{3}{c}{60.8}\tabularnewline
\hline 
SOE-Net (3D)  & \multicolumn{3}{c|}{\textbf{95.6}} & \multicolumn{3}{c|}{\textbf{91.7}} & \multicolumn{3}{c}{\textbf{91.0}}\tabularnewline
\hline 
\end{tabular}
\par\end{centering}
\begin{centering}
} 
\par\end{centering}
\caption{\label{3D-VS-2D}Benefits of 3D vs. 2D filtering.}
\vspace{-6pt}
\end{table}

\begin{table*}[!]
\begin{centering}
\resizebox{0.9\textwidth}{0.04\textheight}{ 
\par\end{centering}
\begin{centering}
\begin{tabular}{c|ccc|ccc|ccc}
\hline 
 & \multicolumn{3}{c|}{YUVL1} & \multicolumn{3}{c|}{YUVL2} & \multicolumn{3}{c}{YUVL3}\tabularnewline
\hline 
 & Scale 1  & Scale 2  & {[}Scale 1, Scale 2{]}  & Scale 1  & Scale 2  & {[}Scale 1, Scale 2{]}  & Scale 1  & Scale 2  & {[}Scale 1, Scale 2{]}\tabularnewline
\hline 
Layer 1  & 91.1  & 87.5  & 92.9  & 90.2  & 85.3  & 90.9  & 84.6  & 80.8  & 86.2\tabularnewline
Layer 2  & 94.3  & 91.9  & \textbf{95.6}  & 89.0  & 88.2  & \textbf{91.7}  & 86.7  & 86.9  & \textbf{91.0}\tabularnewline
\hline 
\cite{KostaTexture}  & -  & -  & 94.0  & -  & -  & 90.0  & -  & -  & -\tabularnewline
\hline 
\end{tabular}
\par\end{centering}
\begin{centering}
}
\par\end{centering}
\caption{\label{scales-and-layers} Classification accuracy on the YUVL dataset
using SOE-Net with multiple layers and scales.}
\vspace{-15pt}
\end{table*}

\vspace{-5pt}
 \textbf{Multiple layers and scales.} Table~\ref{scales-and-layers}
documents the benefits of multiscale filtering, \eqref{eq:convolution}, at each level of the proposed
network as well as those of multiple layers, \eqref{eq:recurrence}. The results show that
addition of multiple layers and scales consistently improve classification
accuracy, with an increase ranging from $\sim2\%$ to $\sim6\%$ in
going from a single layer and scale to multiple layers and scales.
Significantly, the combined multiscale, multilayer results also outperform
the best results previously presented on this dataset of $94\%$ and
$90\%$ for YUVL1 and YUVL2, respectively \cite{KostaTexture}. (There
are no previously reported results on YUVL3.) These results highlight
the pivotal role of the recurrent connection in the proposed network
that allows it to decompose the input signal to reveal novel information
across scales and layers.

Notably, the network instantiation applied to YUVL is limited to 
2 layers and scales due to the small spatiotemporal extent of some
sequences in the dataset; see Sec.~\ref{subsec:Implementation-Details}
for how dataset extent determines layers and scales. Based on these results, all SOE-Net results presented in the remainder of this paper are based on feature vectors formed through the concatenation
of the last layer's output at all scales.

\textbf{Two path rectification.} Next, the advantage of the proposed
two path rectification, \eqref{eq:rectification}, is evaluated. Comparison is made to two alternative
rectification approaches: 1) full wave rectification, where the positive
and negative signals are combined as its input signal is simply pointwise
squared; 2) ReLU rectification, where only the positive part of its
input signal is retained. Table~\ref{two_path-VS-positive-VS-fullwave-1}
shows the benefit of the proposed rectification approach. It is seen
that not only is there value of not neglecting one signal component
(\ie full-wave outperforms ReLU), but moreover additional benefit
comes from keeping the positive and negative components separate,
which suggest their complementarity. 

\vspace{-10pt}
\begin{table}[H]
\begin{centering}
\resizebox{0.8\columnwidth}{0.03\textheight}{ 
\par\end{centering}
\begin{centering}
\begin{tabular}{c|c|c|c}
\hline 
 & \multicolumn{1}{c|}{YUVL1} & \multicolumn{1}{c|}{YUVL2} & \multicolumn{1}{c}{YUVL3}\tabularnewline
\hline 
Full Wave Rectification  & 94.2  & 90.3  & 89.3\tabularnewline
\hline 
Positive Path (ReLU)  & 94.2  & 89.4  & 88.4\tabularnewline
\hline 
Two Paths  & \textbf{95.6}  & \textbf{91.7}  & \textbf{91.0}\tabularnewline
\hline 
\end{tabular}
\par\end{centering}
\begin{centering}
} 
\par\end{centering}
\caption{\label{two_path-VS-positive-VS-fullwave-1}Benefits of the proposed
two path rectification approach.}
\vspace{-10pt}
\end{table}

\textbf{Normalization.} Normalization, \eqref{eq:normalization}, serves to increase invariance to contrast changes.  To document this advantage,  an instantiation of SOE-Net without
the normalization block is compared. The results in Table~\ref{norm-VS-nonorm} clearly demonstrate the usefulness of this step.

\vspace{-10pt}
\begin{table}[H]
\begin{centering}
\resizebox{0.8\columnwidth}{0.02\textheight}{ 
\par\end{centering}
\begin{centering}
\begin{tabular}{c|c|c|c}
\hline 
 & \multicolumn{1}{c|}{YUVL1} & \multicolumn{1}{c|}{YUVL2} & \multicolumn{1}{c}{YUVL3}\tabularnewline
\hline 
No Normalization & 90.6 & 87.2 & 82.8\tabularnewline
\hline 
With Normalization & \textbf{95.6}  & \textbf{91.7}  & \textbf{91.0}\tabularnewline
\hline 
\end{tabular}
\par\end{centering}
\begin{centering}
} 
\par\end{centering}
\caption{\label{norm-VS-nonorm}Benefits of the normalization step.}
\vspace{-10pt}
\end{table}

\textbf{Spatiotemporal pooling.} The theoretical basis for use of a Gaussian
filter in spatiotemporal pooling was given in Sec.~\ref{sub:pooling}.
Here, this choice is validated through comparison to two alternative
pooling approaches \cite{Jarret2009}: 1) the Gaussian filter is replaced with a simple
boxcar filter; 2) the more widely used max pooling is considered. Table~\ref{two_path-VS-positive-VS-fullwave-1-1}
shows the benefit of the proposed spatiotemporal pooling approach
that outperforms other pooling strategies by at least $4\%$. 

\vspace{-10pt}
\begin{table}[H]
\begin{centering}
\resizebox{0.8\columnwidth}{0.03\textheight}{ 
\par\end{centering}
\begin{centering}
\begin{tabular}{c|c|c|c}
\hline 
 & \multicolumn{1}{c|}{YUVL1} & \multicolumn{1}{c|}{YUVL2} & \multicolumn{1}{c}{YUVL3}\tabularnewline
\hline 
boxcar filter  & 91.7 & 87.3 & 87.0\tabularnewline
\hline 
max pooling & 91.6 & 87.4 & 85.7\tabularnewline
\hline 
Gaussian filter  & \textbf{95.6} & \textbf{91.7} & \textbf{91.0 }\tabularnewline
\hline 
\end{tabular}
\par\end{centering}
\begin{centering}
} 
\par\end{centering}
\caption{\label{two_path-VS-positive-VS-fullwave-1-1}Benefits of the used
spatiotemporal pooling approach.}
\vspace{-10pt}
\end{table}

\textbf{Cross-channel pooling.} SOE-Net's novel cross-channel
(CC) pooling plays a pivotal role in keeping the dimensionality manageable. Table~\ref{withCC-VS-noCC} compares the size of the feature vectors with and without CC-pooling for $|\sigma|=1$. Note that dimensionality reduction does not occur until layer 3, as final feature vector output occurs prior to where CC-pooling would apply and is vacuous at level 1; see Sec.~\ref{sub:pooling}.
To examine the impact of such striking dimensionality reduction on classification, SOE-Net is compared with and without CC-pooling. Here, a dataset with spatiotemporal size big enough
to support 3 layers is needed, as it is the point where dimensionality reduction becomes apparent. Also, comparing beyond layer 3 is not computationally feasible due
to dimensionality explosion without CC-pooling, even though it is reasonable with CC-pooling, as done in Secs.~\ref{subsec:Comparison-to-C3D} and \ref{subsec:Comparison-to-DT-SOA}. Since the spatial dimensions of YUVL only support up to layer 2, two variations of  Dyntex (Beta and Dyntex\_35) are used here. Significantly, on both datasets accuracy of SOE-Net with vs. without CC-pooling is comparable, \ie $97.7\%$ vs. $96.8\%$ on Dyntex\_35 and $95.7\%$ vs. $95.1\%$ on Beta, although the feature vectors extracted using CC-pooling are an order of magnitude smaller. These results show that CC-pooling keeps network size manageable while maintaining high discriminating power.\vspace{-8pt}
\begin{table}[H]
\begin{centering}
\resizebox{0.8\columnwidth}{0.025\textheight}{ 
\par\end{centering}
\begin{centering}
\begin{tabular}{c|c|c|c|c|c}
\hline 
 & \multicolumn{1}{c|}{L1} & \multicolumn{1}{c|}{L2} & \multicolumn{1}{c|}{L3} & \multicolumn{1}{c|}{L4} & \multicolumn{1}{c}{L5}\tabularnewline
\hline 
SOE-Net (a)  & 20 & 400 & 8000 & 160000 & 3200000\tabularnewline
\hline 
SOE-Net (b) & \textbf{20} & \textbf{400} & \textbf{800} & \textbf{1600} & \textbf{3200}\tabularnewline
\hline 
\end{tabular}
\par\end{centering}
\begin{centering}
}
\par\end{centering}
\caption{\label{withCC-VS-noCC} Feature dimensions (a) without vs. (b) with
CC-pooling.}
\vspace{-10pt}
\end{table}

\subsection{Comparison to a learned 3D ConvNet}\label{subsec:Comparison-to-C3D}

\vspace{-5pt}
\begin{table*}
\begin{centering}
\resizebox{0.75\textwidth}{0.025\textheight}{ 
\par\end{centering}
\begin{centering}
\begin{tabular}{c|c|c|c|c|c|c|c}
\hline 
 & YUVL1  & YUVL2  & YUVL3  & Alpha  & Beta  & Gamma  & Dyntex\_35 \tabularnewline
\hline 
C3D \cite{Tran2015}  & 88.0  & 89.8  & 85.5  & 100  & 96.3  & 95.0  & 96.3 \tabularnewline
\hline 
SOE-Net  & \textbf{95.6}  & \textbf{91.7}  & \textbf{91.0}  & 98.3  & \textbf{96.9}  & 93.6  & \textbf{97.7}\tabularnewline
\hline 
\end{tabular}
\par\end{centering}
\begin{centering}
}
\par\end{centering}
\caption{\label{tab:SOEvsC3D} Comparison of SOE-Net features versus C3D features.}
\vspace{-7pt}
\end{table*}

\begin{table*}
\begin{centering}
\resizebox{0.9\textwidth}{0.08\textheight}{ 
\par\end{centering}
\begin{centering}
\begin{tabular}{c|c|c|c|c|c|c|c|c|c|c}
\hline 
\multicolumn{2}{c|}{} & \multicolumn{2}{c|}{Alpha} & \multicolumn{2}{c|}{Beta} & \multicolumn{2}{c|}{Gamma} & \multicolumn{2}{c|}{Dyntex\_35} & \multicolumn{1}{c}{Dyntex++}\tabularnewline
\cline{3-11} 
\multicolumn{2}{c|}{Method} & SVM  & NCC  & SVM  & NCC  & SVM  & NCC  & NN  & NCC  & SVM\tabularnewline
\hline 
\multirow{4}{*}{Learning-based} & \cite{Ghanem2010}  & -  & -  & -  & -  & -  & -  & -  & -  & 63.7\tabularnewline
 & \cite{Mumtaz2013}  & -  & -  &  & -  & -  & -  & 98.6  & -  & -\tabularnewline
 & \cite{Harandi2013}  & -  & -  &  & -  & -  & -  & -  & -  & 92.8\tabularnewline
 & \cite{QuanHJ15}  & 87.8  & 86.6  & 76.7  & 69.0  & 74.8  & 64.2  & \textbf{99.0}  & \textbf{97.8}  & \textbf{94.7}\tabularnewline
\hline 
\multirow{5}{*}{Hand-crafted} & \cite{Xu2011}  & 84.9  & 83.6  & 76.5  & 65.2  & 74.5  & 60.8  & -  & 97.6  & 89.9\tabularnewline
 & \cite{Xu2012}  & 82.8  & -  & 75.4  & -  & 73.5  & -  & -  & 96.7  & 89.2\tabularnewline
 & \cite{Ji2013}  & -  & -  & -  & -  & -  & -  & -  & 96.5  & 88.8\tabularnewline
 & \cite{Zhao2007}  & 83.3  & -  & 73.4  & -  & 72.0  & -  & -  & 97.1  & 89.8\tabularnewline
 & \cite{Dubois2013}  & -  & 85.0  & -  & 67.0  & -  & 63.0  & -  & -  & -\tabularnewline
\hline 
 & SOE-NET %- HOLD OUT 
  & \textbf{96.7}  & \textbf{96.7}  & \textbf{95.7}  & \textbf{\textcolor{black}{86.4}}  & \textbf{92.2}  & \textbf{\textcolor{black}{80.3}}  & \textcolor{black}{97.7}  & \textcolor{black}{93.1}  & 94.4\tabularnewline
\hline 
\end{tabular}
\par\end{centering}
\begin{centering}
} 
\par\end{centering}
\caption{\label{DT-COMPARISON}Comparison to state-of-the-art methods on Dynamic
Texture recognition.}
\vspace{-15pt}
\end{table*}

It is interesting to compare the performance of SOE-Net to a learning based 3D ConvNet. For this comparison C3D \cite{Tran2015} is used
for 3 main reasons. 1) Currently, C3D is the only ConvNet trained
end-to-end using 3D filters only without any pre-processing of the
input volumes (\eg to extract optical flow). This architecture makes
it the most similar trained network to SOE-Net that also relies on
3D convolutions on the raw input volumes. 2) This pre-trained network
has shown state-of-the-art performance, without any fine tuning,
on a variety of tasks, including action recognition, object recognition and dynamic scene classification, which is very similar to dynamic
texture recognition. Indeed, C3D is advocated as a general feature
extractor for video analysis tasks without fine tuning (see Sec. 3.3
in \cite{Tran2015}). 3) It is not feasible to fine tune a network
with any of the available DT datasets due to their very
limited size. So, alternative ConvNets that require such data for
fine tuning prior to testing cannot be compared in a meaningful fashion.

C3D features are extracted (FC-6 activations as specified in
\cite{Tran2015}) for all versions of the YUVL and Dyntex,
except Dyntex++, which is discarded as the small size of its videos
($50\times50\times50$) precludes application of C3D. For experiments
with the considered Dyntex versions, the relatively large size of
the sequences makes it possible to push SOE-Net to 3 layers on Dyntex\_35
and to 5 layers on Alpha, Beta, Gamma. The results  in Table~\ref{tab:SOEvsC3D} show
that SOE-Net outperformed C3D on the majority of cases (5 of 7). The
larger performance gaps between SOE-Net and C3D on the YUVL dataset are of particular note, especially given that SOE-Net
uses only 2 layers on YUVL.

Significantly, design of YUVL was motivated by interest in building
a true {\em dynamic} texture dataset. So, it groups patterns based
on their dynamics, rather than their appearance. Thus, relative performance on YUVL suggests that SOE-Net is more
able to capitalize on dynamic information than C3D.
Further, SOE-Net either outperforms or is on par with C3D on datasets
that group dynamic textures with more emphasis on visual appearance,
\ie, Dyntex, suggesting that SOE-Net is able to capitalize on appearance
information as well. Overall, these results show that SOE-Net can
capture rich, discriminative information, be it dynamic or static
appearance, without reliance on extensive and costly training.

\vspace{-5pt}

\subsection{Dynamic texture recognition state-of-the-art\label{subsec:Comparison-to-DT-SOA}}

\vspace{-2pt}
 Comparison now is made to state-of-the-art approaches designed for
dynamic texture recognition using the currently most widely used DT datasets (\ie various breakdowns of Dyntex, as YUVL has received
notably less attention). In particular, SOE-Net is compared to 4 learning
based \cite{Ghanem2010,Mumtaz2013,Harandi2013,QuanHJ15} and 5 hand-crafted
\cite{Xu2011,Xu2012,Zhao2007,Ji2013,Dubois2013} DT descriptors. Following
previous research using Dyntex \cite{Xu2011,Xu2012,Zhao2007,Ji2013,Dubois2013,QuanHJ15},
evaluation is performed variously using SVM, Nearest Class Center
(NCC) and Nearest Neighbor (NN) classifiers. For SVM, the same protocol
used in previous research is followed, whereby $50\%$ of samples per
category are used for training and the rest for testing. 

Table~\ref{DT-COMPARISON} shows that SOE-Net outperforms all other
approaches on the Alpha, Beta and Gamma datasets, with sizable performance
gaps between 
$\approx9-19\%$ using SVM. Using NCC, SOE-Net outperforms all other methods by at least $10\%$. Overall,
these results speak decisively for the high discriminative power of
SOE-Net's descriptors. 

Similarly, on Dyntex++ SOE-Net extracts stronger features than all
hand-crafted methods, achieving an accuracy that is $4.5\%$ higher
than previous state-of-the-art approach \cite{Xu2011}. Also, under
SVM, SOE-Net is on par with state-of-the-art learning based method,
with only marginal difference ($\approx0.3\%$ ). Moreover, using a NN classifier SOE-Net
outperforms all other methods, with an accuracy of $95.6\%$ (results not shown in table, as not reported
by others). These results again confirm the discriminative power of
the proposed representation. Notably, due to the extremely small size
of the Dyntex++ videos, SOE-Net was restricted to only a single scale
and 2 layers on this particular dataset. 

\begin{figure}
\begin{centering}
\resizebox{\columnwidth}{!}{
\par\end{centering}
\begin{centering}
\begin{tabular}{cccc}
\hline 
steam1-clup1  & steam2-clup2  & waterboil2-clup & curly-hair\tabularnewline
\hline 
\textbf{(Q)} & \textbf{(N)} & \textbf{(Q)} & \textbf{(N)}\tabularnewline
\hline 
\end{tabular}
\par\end{centering}
\begin{centering}
}
\par\end{centering}
\vspace{-1pt}

\begin{centering}
\resizebox{\columnwidth}{!}{
\par\end{centering}
\begin{centering}
\begin{tabular}{cc||cc}
\includegraphics{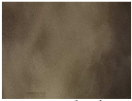}  & \includegraphics{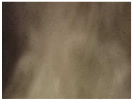}  & \includegraphics{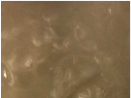} & \includegraphics{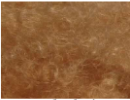}\tabularnewline
\end{tabular}
\par\end{centering}
\begin{centering}
}
\par\end{centering}
\caption{\label{missclass}Misclassification examples on Dyntex\_35 using NCC.
\textbf{(Q)} is the query class. (\textbf{N}) is the nearest class
in the database.}

\vspace{-16pt}
\end{figure}

Finally, on Dyntex\_35 SOE-Net performs slightly worse than state-of-the-art
using the NN classifier (a difference of $\approx$1.3\%); however,
the difference is greater using the NCC classifier. Interestingly,
closer examination of these results reveals that confusions typically
occur between slightly different views of the same physical process,
which naturally yield visually very similar dynamic textures. Other
confusions arise from different physical processes, which happen to
yield similar visual appearances. Figure~\ref{missclass} shows examples. Overall, the fact that these are the types
of confusions made by the network suggests an ability to generalize
across viewpoint, which arguably is more important than
ability to make fine grained distinctions between viewpoints during
texture classification as well as other recognition tasks. 

\vspace{-4pt}

%------------------------------------------------------------------------- 

\vspace{-5pt}
\section{Conclusion\label{sec:conclusionl} }

\vspace{-5pt}
 This paper presented a novel hierarchical spatiotemporal network  based on three key ideas. First, a multilayer repeated filtering architecture
is employed. Second, design decisions have been theoretically motivated
without relying on learning or other empirically driven decisions.
Third, in addition to adding insight into convolution, rectification,
normalization and spatiotemporal pooling,
a novel cross-channel pooling has been introduced that keeps the representation compact
while maintaining representational clarity. The repeated filtering architecture
and theory driven design makes the representation understandable in
terms of multiorientation, multiscale properties. Further, by eschewing
learning, the approach does not rely on training data. Finally, the
benefits of SOE-Net were shown on dynamic texture recognition, where
it extends the state-of-the-art.

\newpage
\vspace{-2pt}
{\small{}{}  \bibliographystyle{ieee}
\bibliography{mybibtex}
 }{\small \par}

\end{document}